\documentclass[10pt, a4paper]{article}
\usepackage{lrec}
\usepackage{graphicx}
\usepackage{tabularx}
\usepackage{soul}

\usepackage{epstopdf}
\usepackage[utf8]{inputenc}

\usepackage{hyperref}
\usepackage{xstring}
\usepackage{amsmath}
\usepackage{amssymb}
\usepackage{color}

\title{Diversity, Density, and Homogeneity: \\ Quantitative Characteristic Metrics for Text Collections}

\name{Yi-An Lai$^{\dagger1}$\thanks{$^{\dagger}$These authors contributed equally.}, Xuan Zhu$^{\dagger1}$, Yi Zhang$^{1}$, Mona Diab$^{*2}$\thanks{$^{*}$Work done as a Principal Scientist at AWS AI.}}

\address{$^{1}$Amazon AWS AI \\
         $^{2}$The George Washington University \\
         \{yianl, zhuxuan, yizhngn\}@amazon.com, mtdiab@gwu.edu\\}

\abstract{
Summarizing data samples by quantitative measures has a long history, with descriptive statistics being a case in point.
However, as natural language processing methods flourish, there are still insufficient characteristic metrics to describe a collection of texts in terms of the words, sentences, or paragraphs they comprise.
In this work, we propose metrics of diversity, density, and homogeneity  that quantitatively measure the dispersion, sparsity, and uniformity of a text collection.
We conduct a series of simulations to verify that each metric holds desired properties and resonates with human intuitions.
Experiments on real-world datasets demonstrate that the proposed characteristic metrics are highly correlated with text classification performance of a renowned model, BERT, which could inspire future applications.\\ \newline 
\Keywords{diversity, density, homogeneity, characteristics, text classification, quantitative measures} }

\begin{document}

\maketitleabstract

\section{Introduction}
\bigskip
Characteristic metrics are a set of unsupervised measures that quantitatively describe or summarize the properties of a data collection.
These metrics generally do not use ground-truth labels and only measure the intrinsic characteristics of data.
The most prominent example is descriptive statistics that summarizes a data collection by a group of unsupervised measures such as mean or median for central tendency, variance or minimum-maximum for dispersion, skewness for symmetry, and kurtosis for heavy-tailed analysis.

In recent years, text classification, a category of Natural Language Processing (NLP) tasks, has drawn much attention \cite{zhang2015character,joulin2016bag,howard2018universal} for its wide-ranging real-world applications such as fake news detection \cite{shu2017fake}, document classification \cite{yang2016hierarchical}, and spoken language understanding (SLU) \cite{gupta2019simple,gupta2019casa,zhang2018joint}, 
a core task of conversational assistants like Amazon Alexa or Google Assistant.

However, there are still insufficient characteristic metrics to describe a collection of texts.
Unlike numeric or categorical data, simple descriptive statistics alone such as word counts and vocabulary size are difficult to capture the syntactic and semantic properties of a text collection.

In this work, we propose a set of characteristic metrics: \emph{diversity}, \emph{density}, and \emph{homogeneity} to quantitatively summarize a collection of texts where the unit of texts could be a phrase, sentence, or paragraph.
A text collection is first mapped into a high-dimensional embedding space. 
Our characteristic metrics are then computed to measure the dispersion, sparsity, and uniformity of the distribution.
Based on the choice of embedding methods, these characteristic metrics can help understand the properties of a text collection from different linguistic perspectives, for example, lexical diversity, syntactic variation, and semantic homogeneity.
Our proposed diversity, density, and homogeneity metrics extract hard-to-visualize quantitative insight for a better understanding and comparison between text collections.

To verify the effectiveness of proposed characteristic metrics, we first conduct a series of simulation experiments that cover various scenarios in two-dimensional as well as high-dimensional vector spaces.
The results show that our proposed quantitative characteristic metrics exhibit several desirable and intuitive properties such as robustness and linear sensitivity of the diversity metric with respect to random down-sampling.
Besides, we investigate the relationship between the characteristic metrics and the performance of a renowned model, BERT \cite{devlin2018bert}, on the text classification task using two public benchmark datasets.
Our results demonstrate that there are high correlations between text classification model performance and the characteristic metrics, which shows the efficacy of our proposed metrics. 

\section{Related Work}
\bigskip
A building block of characteristic metrics for text collections is the language representation method.
A classic way to represent a sentence or a paragraph is n-gram, with dimension equals to the size of vocabulary.
More advanced methods learn a relatively low dimensional latent space that represents each word or token as a continuous semantic vector such as word2vec \cite{mikolov2013distributed}, GloVe \cite{pennington2014glove}, and fastText \cite{mikolov2017advances}.
These methods have been widely adopted with consistent performance improvements on many NLP tasks.
Also, there has been extensive research on representing a whole sentence as a vector such as a plain or weighted average of word vectors \cite{arora2016simple}, skip-thought vectors \cite{kiros2015skip}, and self-attentive sentence encoders \cite{lin2017structured}.

More recently, there is a paradigm shift from non-contextualized word embeddings to self-supervised language model (LM) pretraining.
Language encoders are pretrained on a large text corpus using a LM-based objective and then re-used for other NLP tasks in a transfer learning manner.
These methods can produce contextualized word representations, which have proven to be effective for significantly improving many NLP tasks.
Among the most popular approaches are ULMFiT \cite{howard2018universal}, ELMo \cite{peters2018deep}, OpenAI GPT \cite{radford2018improving}, and BERT \cite{devlin2018bert}.
In this work, we adopt BERT, a transformer-based technique for NLP pretraining, as the backbone to embed a sentence or a paragraph into a representation vector.

Another stream of related works is the evaluation metrics for cluster analysis.
As measuring property or quality of outputs from a clustering algorithm is difficult, human judgment with cluster visualization tools \cite{kwon2017clustervision,kessler2017scattertext} are often used.
There are unsupervised metrics to measure the quality of a clustering result such as the Calinski-Harabasz score \cite{calinski1974dendrite}, the Davies-Bouldin index \cite{davies1979cluster}, and the Silhouette coefficients \cite{rousseeuw1987silhouettes}.
Complementary to these works that model cross-cluster similarities or relationships, our proposed diversity, density and homogeneity metrics focus on the characteristics of each single cluster, i.e., intra cluster rather than inter cluster relationships.

\section{Proposed Characteristic Metrics}
\bigskip
We introduce our proposed \emph{diversity}, \emph{density}, and \emph{homogeneity} metrics with their detailed formulations and key intuitions.

Our first assumption is, for classification, high-quality training data entail that examples of one class are as differentiable and distinct as possible from another class.
From a fine-grained and intra-class perspective, a robust text cluster should be diverse in syntax, which is captured by \emph{diversity}.
And each example should reflect a sufficient signature of the class to which it belongs, that is, each example is representative and contains certain salient features of the class. 
We define a \emph{density} metric to account for this aspect.
On top of that, examples should also be semantically similar and coherent among each other within a cluster, where \emph{homogeneity} comes in play.

The more subtle intuition emerges from the inter-class viewpoint.
When there are two or more class labels in a text collection, in an ideal scenario, we would expect the homogeneity to be monotonically decreasing. 
Potentially, the diversity is increasing with respect to the number of classes since text clusters should be as distinct and separate as possible from one another.
If there is a significant ambiguity between classes, the behavior of the proposed metrics and a possible new metric as a inter-class confusability measurement remain for future work.

In practice, the input is a collection of texts $\{x_1, x_2, ..., x_m\}$, where $x_i$ is a sequence of tokens $x_{i1}, x_{i2}, ..., x_{il}$ denoting a phrase, a sentence, or a paragraph.
An embedding method $\mathcal{E}$ then transforms $x_i$ into a vector $\mathcal{E}(x_i)=e_i$ and the characteristic metrics are computed with the embedding vectors. For example, 
\begin{equation}
M_{diversity} = f_{diversity}(\{e_1, e_2, ..., e_m\}).
\end{equation}
Note that these embedding vectors often lie in a high-dimensional space, e.g. commonly over $300$ dimensions. 
This motivates our design of characteristic metrics to be sensitive to text collections of different properties while being robust to the curse of dimensionality.

We then assume a set of clusters created over the generated embedding vectors. In classification tasks, the embeddings pertaining to members of a class form a cluster, i.e., in a supervised setting. 
In an unsupervised setting, we may apply a clustering algorithm to the embeddings. 
It is worth noting that, in general, the metrics are independent of the assumed underlying grouping method.

\subsection{Diversity}
\bigskip
Embedding vectors of a given group of texts $\{e_1, ..., e_m\}$ can be treated as a cluster in the high-dimensional embedding space.
We propose a diversity metric to estimate the cluster's dispersion or spreadness via a generalized sense of the radius.

Specifically, if a cluster is distributed as a multi-variate Gaussian with a diagonal covariance matrix $\Sigma$, the shape of an isocontour will be an axis-aligned ellipsoid in $\mathbb{R}^{H}$.
Such isocontours can be described as:

\begin{equation}
(\boldsymbol{x} - \boldsymbol{\mu})^{T}\boldsymbol{\Sigma}^{-1}(\boldsymbol{x} - \boldsymbol{\mu}) = \sum^{H}_{j=1}{\frac{(x_j - \mu_j)^2}{\sigma^2_{j}}} = c^2,
\end{equation}
where $\boldsymbol{x}$ are all possible points in $\mathbb{R}^{H}$ on an isocontour, $c$ is a constant, $\boldsymbol{\mu}$ is a given mean vector with $\mu_j$ being the value along $j$-th axis, and $\sigma^2_j$ is the variance of the $j$-th axis.

We leverage the geometric interpretation of this formulation and treat the square root of variance, i.e., standard deviation, $\sqrt{\sigma^2_j}$ as the radius $r_j$ of the ellipsoid along the $j$-th axis.
The diversity metric is then defined as the geometric mean of radii across all axes:

\begin{equation}
\begin{split}
M_{diversity}  &= (r_1 \cdot r_2 \cdot... \cdot r_{H})^{\frac{1}{H}}\\
 &= (\sqrt{\sigma^2_1} \cdot ... \cdot \sqrt{\sigma^2_H})^{\frac{1}{H}} = \sqrt[H]{\prod^{H}_{i=i}{\sigma_i}},
\end{split}
\end{equation}
where $\sigma_i$ is the standard deviation or square root of the variance along the $i$-th axis.

In practice, to compute a diversity metric, we first calculate the standard deviation of embedding vectors along each dimension and take the geometric mean of all calculated values.
Note that as the geometric mean acts as a dimensionality normalization, it makes the diversity metric work well in high-dimensional embedding spaces such as BERT.

\subsection{Density}
\bigskip
Another interesting characteristic is the sparsity of the text embedding cluster.
The density metric is proposed to estimate the number of samples that falls within a unit of volume in an embedding space.

Following the assumption mentioned above, a straight-forward definition of the volume can be written as:

\begin{equation}
(r_1 \cdot ... \cdot r_H) = (\sqrt{\sigma^2_1} \cdot ... \cdot \sqrt{\sigma^2_H}) = \prod^{H}_{i=i}{\sigma_i},
\end{equation}
up to a constant factor.
However, when the dimension goes higher, this formulation easily produces exploding or vanishing density values, i.e., goes to infinity or zero.

To accommodate the impact of high-dimensionality, we impose a dimension normalization. 
Specifically, we introduce a notion of effective axes, which assumes most variance can be explained or captured in a sub-space of a dimension $\sqrt{H}$.
We group all the axes in this sub-space together and compute the geometric mean of their radii as the effective radius.
The dimension-normalized volume is then formulated as:

\begin{equation}
\begin{split}
volume & = (r_1 \cdot ... \cdot r_{\sqrt{H}})^{\frac{1}{\sqrt{H}}} ... (r_{H - \sqrt{H} + 1} \cdot ...  \cdot r_{H})^{\frac{1}{\sqrt{H}}}\\
 & =(r_1 \cdot ... \cdot r_H)^{\frac{1}{\sqrt{H}}} = (\prod^{H}_{i=i}{\sigma_i})^{\frac{1}{\sqrt{H}}}
\end{split}
\end{equation}

Given a set of embedding vectors $\{e_1, ..., e_m\}$, we define the density metric as:

\begin{equation}
M_{density} = \frac{m}{(\prod^{H}_{i=i}{\sigma_i})^{\frac{1}{\sqrt{H}}}}
\end{equation}

In practice, the computed density metric values often follow a heavy-tailed distribution, thus sometimes its $\log$ value is reported and denoted as $density (log\-scale)$.

\subsection{Homogeneity}
\bigskip
The homogeneity metric is proposed to summarize the uniformity of a cluster distribution.
That is, how uniformly the embedding vectors of the samples in a group of texts are distributed in the embedding space.
We propose to quantitatively describe homogeneity by building a fully-connected, edge-weighted network, which can be modeled by a Markov chain model.
A Markov chain's entropy rate is calculated and normalized to be in $[0, 1]$ range by dividing by the entropy's theoretical upper bound.
This output value is defined as the homogeneity metric detailed as follows:

To construct a fully-connected network from the embedding vectors $\{e_1, ..., e_m\}$, we compute their pairwise distances as edge weights, an idea similar to AttriRank \cite{hsu2017unsupervised}\footnote{\url{https://github.com/ntumslab/AttriRank/blob/master/attrirank.pdf}}.
As the Euclidean distance is not a good metric in high-dimensions, we normalize the distance by adding a power $\log(n\_dim)$.
We then define a Markov chain model with the weight of $edge(i, j)$ being

\begin{equation}
    weight(i, j) = \left( \sqrt{(\boldsymbol{e_i} - \boldsymbol{e_j}) \cdot (\boldsymbol{e_i} - \boldsymbol{e_j})} \right) ^{\log(H)}
\end{equation}

and the conditional probability of transition from $i$ to $j$ can be written as

\begin{equation}
    p(i \rightarrow j) = \frac{weight(i, j)}{\sum_{k}{weight(i, k)}}.
\end{equation}

All the transition probabilities $p(i \rightarrow j)$ are from the transition matrix of a Markov chain.
An entropy of this Markov chain can be calculated\footnote{\url{https://en.wikipedia.org/wiki/Entropy_rate}} as

\begin{equation}
    entropy = - \sum_{ij}{\nu_{i}\cdot p(i \rightarrow j) \log p(i \rightarrow j)},
\end{equation}

where $\nu_i$ is the stationary distribution of the Markov chain.
As self-transition probability $p(i \rightarrow i)$ is always zero because of zero distance, there are $(m - 1)$ possible destinations and the entropy's theoretical upper bound becomes 

\begin{equation}
    - \sum_{ij, i\neq j}{\left(\frac{1}{m}\right)\cdot \frac{1}{m - 1} \log \frac{1}{m-1}} = \log (m - 1).
\end{equation}

Our proposed homogeneity metric is then normalized into $[0, 1]$ as a uniformity measure:

\begin{equation}
    M_{homogeneity} = \frac{-\sum_{ij}{\nu_{i}\cdot p(i \rightarrow j) \log p(i \rightarrow j)}}{\log (m - 1)}.
\end{equation}

The intuition is that if some samples are close to each other but far from all the others, the calculated entropy decreases to reflect the unbalanced distribution.
In contrast, if each sample can reach other samples within more-or-less the same distances, the calculated entropy as well as the homogeneity measure would be high as it implies the samples could be more uniformly distributed.

\section{Simulations}
\bigskip
To verify that each proposed characteristic metric holds its desirable and intuitive properties, we conduct a series of simulation experiments in $2$-dimensional as well as $768$-dimensional spaces. The latter has the same dimensionality as the output of our chosen embedding method-BERT, in the following Experiments section.

\begin{figure*}
\begin{center}
\includegraphics[width=0.9\textwidth]{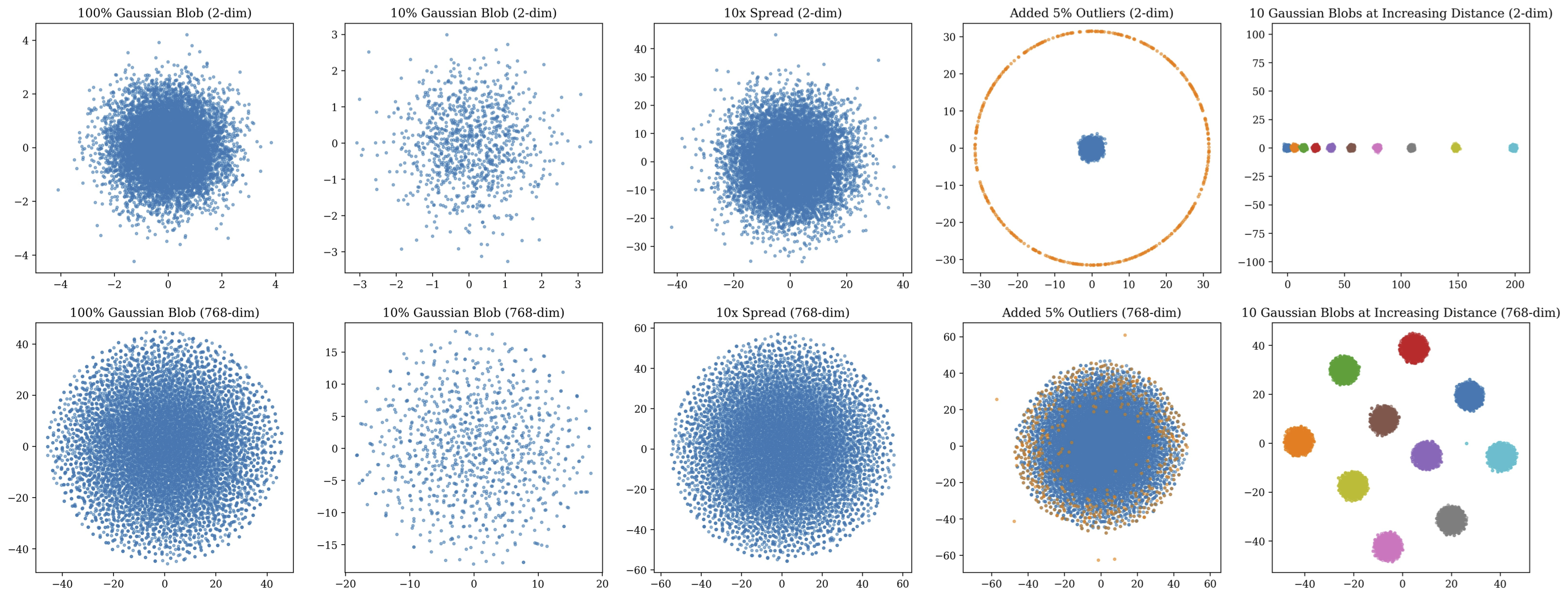} 
\caption{Visualization of the simulations including base setting, down-sampling, varying spreads, adding outliers, and multiple sub-clusters in $2$-dimensional and $768$-dimensional spaces.}
\label{simulations}
\end{center}
\end{figure*}

\subsection{Simulation Setup}
\bigskip
The base simulation setup is a randomly generated isotropic Gaussian blob that contains $10,000$ data points with the standard deviation along each axis to be $1.0$ and is centered around the origin.
All Gaussian blobs are created using \texttt{make\_blobs} function in the \texttt{scikit-learn} package\footnote{\url{https://scikit-learn.org/stable}}.

Four simulation scenarios are used to investigate the behavior of our proposed quantitative characteristic metrics:

\begin{itemize}
    \item Down-sampling: Down-sample the base cluster to be $\{90\%, 80\%, ..., 10\%\}$ of its original size. That is, create Gaussian blobs with $\{9000, ..., 1000\}$ data points;
    \item Varying Spread: Generate Gaussian blobs with standard deviations of each axis to be $\{2.0, 3.0, ..., 10.0\}$;
    \item Outliers: Add $\{50, 100, ..., 500\}$ outlier data points, i.e., $\{0.5\%, ..., 5\%\}$ of the original cluster size, randomly on the surface with a fixed norm or radius;  
    \item Multiple Sub-clusters: Along the $1$th-axis, with $10,000$ data points in total, create $\{1, 2, ..., 10\}$ clusters with equal sample sizes but at increasing distance.
\end{itemize}

For each scenario, we simulate a cluster and compute the characteristic metrics in both $2$-dimensional and $768$-dimensional spaces.
Figure \ref{simulations} visualizes each scenario by t-distributed Stochastic Neighbor Embedding (t-SNE) \cite{maaten2008visualizing}.
The $768$-dimensional simulations are visualized by down-projecting to $50$ dimensions via Principal Component Analysis (PCA) followed by t-SNE.

\begin{figure*}
\begin{center}
\includegraphics[width=0.85\textwidth]{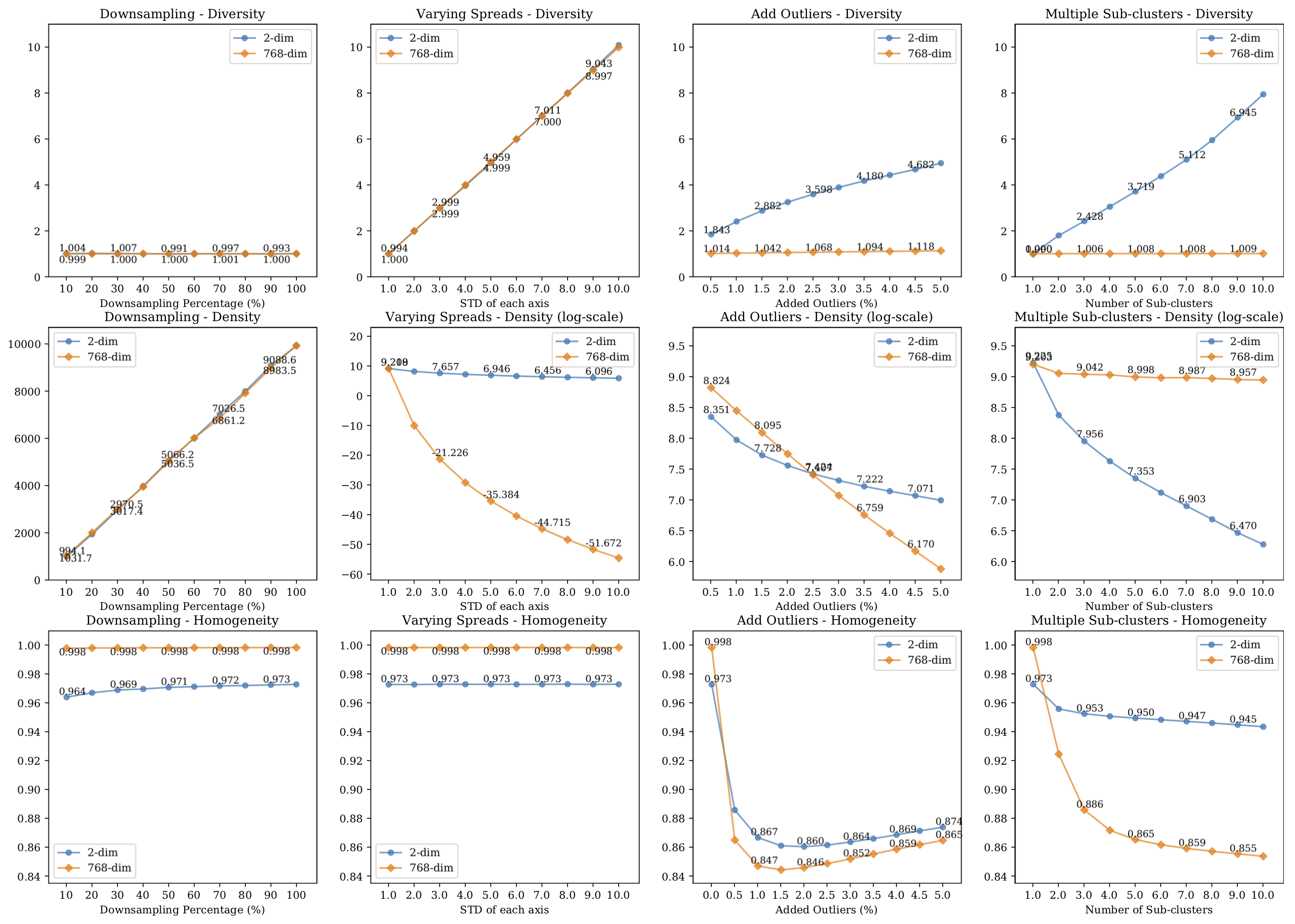} 
\caption{Diversity, density, and homogeneity metric values in each simulation scenario.}
\label{simulation_result}
\end{center}
\end{figure*}

\subsection{Simulation Results}
\bigskip
Figure \ref{simulation_result} summarizes calculated diversity metrics in the first row, density metrics in the second row, and homogeneity metrics in the third row, for all simulation scenarios.

The diversity metric is robust as its values remain almost the same to the down-sampling of an input cluster.
This implies the diversity metric has a desirable property that it is insensitive to the size of inputs.
On the other hand, it shows a linear relationship to varying spreads.
It is another intuitive property for a diversity metric that it grows linearly with increasing dispersion or variance of input data.
With more outliers or more sub-clusters, the diversity metric can also reflect the increasing dispersion of cluster distributions but is less sensitive in high-dimensional spaces.

For the density metrics, it exhibits a linear relationship to the size of inputs when down-sampling, which is desired.
When increasing spreads, the trend of density metrics  corresponds well with human intuition.
Note that the density metrics decrease at a much faster rate in higher-dimensional space as log-scale is used in the figure.
The density metrics also drop when adding outliers or having multiple distant sub-clusters. This makes sense since both scenarios should increase the dispersion of data and thus increase our notion of volume as well.
In multiple sub-cluster scenario, the density metric becomes less sensitive in the higher-dimensional space.
The reason could be that the sub-clusters are distributed only along one axis and thus have a smaller impact on volume in higher-dimensional spaces.

As random down-sampling or increasing variance of each axis should not affect the uniformity of a cluster distribution, we expect the homogeneity metric remains approximately the same values.
And the proposed homogeneity metric indeed demonstrates these ideal properties. 
Interestingly, for outliers, we first saw huge drops of the homogeneity metric but the values go up again slowly when more outliers are added.
This corresponds well with our intuitions that a small number of outliers break the uniformity but more outliers should mean an increase of uniformity because the distribution of added outliers themselves has a high uniformity.

For multiple sub-clusters, as more sub-clusters are presented, the homogeneity should and does decrease as the data are less and less uniformly distributed in the space.

To sum up, from all simulations, our proposed diversity, density, and homogeneity metrics indeed capture the essence or intuition of dispersion, sparsity, and uniformity in a cluster distribution.

\section{Experiments}
\bigskip

The two real-world text classification tasks we used for experiments are sentiment analysis and Spoken Language Understanding (SLU). 

\subsection{Chosen Embedding Method}
\bigskip
BERT is a self-supervised language model pretraining approach based on the Transformer \cite{vaswani2017attention}, a multi-headed self-attention architecture that can produce different representation vectors for the same token in various sequences, i.e., contextual embeddings.

When pretraining, BERT concatenates two sequences as input, with special tokens $[CLS], [SEP], [EOS]$ denoting the start, separation, and end, respectively. 
BERT is then pretrained on a large unlabeled corpus with objective-masked language model (MLM), which randomly masks out tokens, and  the model  predicts the masked tokens. The other classification task is next sentence prediction (NSP). NSP is to predict whether two sequences follow each other in the original text or not.

In this work, we use the pretrained $\text{BERT}_{\text{BASE}}$ which has $12$ layers (L), $12$ self-attention heads (A), and $768$ hidden dimension (H) as the language embedding to compute the proposed data metrics. 
The off-the-shelf pretrained BERT is obtained from GluonNLP\footnote{\url{https://gluon-nlp.mxnet.io/model_zoo/bert/index.html}}.
For each sequence $x_i = (x_{i1}, ..., x_{il})$ with length $l$, BERT takes $[CLS], x_{i1}, ..., x_{il}, [EOS]$ as input and generates embeddings $\{e_{CLS}, e_{i1}, ..., e_{il}, e_{EOS}\}$ at the token level.
To obtain the sequence representation, we use a mean pooling over token embeddings:

\begin{equation}
e_{i} = \frac{(e_{i1} + ... + e_{il})}{l},
\end{equation}

where $e_i \in \mathbb{R}^{H}$. 
A text collection $\{x_1, ..., x_m\}$, i.e., a set of token sequences, is then transformed into a group of H-dimensional vectors $\{e_1, ..., e_m\}$.

We compute each metric as described previously, using three BERT layers L1, L6, and L12 as the embedding space, respectively. 
The calculated metric values are averaged over layers for each class and averaged over classes weighted by class size as the final value for a dataset.

\begin{table*}[ht]
\begin{center}
\begin{tabular}{|l|l|l|l|l|l|}

      \hline
      Down-Sampling	to & Training Set Size & Accuracy & Diversity & Density & Homogeneity\\
      \hline\hline
      100\%	& 67,350 & 0.9266 & 0.292 & 44.487 & 0.928\\
      90\%	& 60,615 & 0.9323 & 0.292 & 44.367 & 0.927\\
      80\%	& 53,880 & 0.9260 & 0.292 & 44.224 & 0.927\\
      70\%	& 47,146 & 0.9266 & 0.292 & 44.071 & 0.925\\
      60\%	& 40,411 & 0.9312 & 0.292 & 43.928 & 0.924\\
      50\%	& 33,676 & 0.9300 & 0.292 & 43.672 & 0.922\\
      40\%	& 26,941 & 0.9243 & 0.292 & 43.384 & 0.919\\
      30\%	& 20,206 & 0.9300 & 0.292 & 43.148 & 0.917\\
      20\%	& 13,471 & \bf{0.9174} & 0.293 & \bf{42.733} & \bf{0.914}\\
      10\%	& 6,736 & \bf{0.9071} & 0.294 & \bf{41.972} & \bf{0.908}\\
      \hline

\end{tabular}
\caption{The experimental results of diversity, density, and homogeneity metrics with classification accuracy on the SST-2 dataset.}
\label{sst2}
\end{center}
\end{table*}


\begin{table*}[ht]
\begin{center}
\begin{tabular}{|l|l|l|l|l|l|l|}

      \hline
      Down-Sampling to & Training Set Size & IC Accuracy (\%) & SL F1 (\%) & Diversity & Density & Homogeneity\\
      \hline\hline
      100\%	& 13,084 & 98.71 & 96.06 & 0.215 & 48.291 & 0.950\\
      90\%	& 11,773 & 98.57 & 95.79 & 0.215 & 48.199 & 0.949\\
      80\%	& 10,465 & 99.00 & 95.55 & 0.215 & 48.109 & 0.949\\
      70\%	& 9,157 & 99.14 & 95.13 & 0.215 & 47.996 & 0.948\\
      60\%	& 7,848 & 98.71 & 95.02 & 0.215 & 47.751 & 0.948\\
      50\%	& 6,541 & 98.86 & 94.38 & 0.215 & 47.660 & 0.945\\
      40\%	& 5,231 & 99.00 & 94.74 & 0.214 & 47.449 & 0.944\\
      30\%	& 3,922 & 98.57 & 93.74 & 0.215 & 47.090 & 0.941\\
      20\%	& 2,614 & \bf{96.42} & \bf{92.63} & 0.214 & \bf{46.877} & \bf{0.939}\\
      10\%  & 1,306 & \bf{87.20} & \bf{89.12} & 0.214 & \bf{46.158} & \bf{0.929}\\
      \hline

\end{tabular}
\caption{The experimental results of diversity, density, and homogeneity metrics with intent classification (IC) accuracy and slot labeling (SL) F1 scores on the Snips dataset. Experimental setup is the same as that in Table \ref{sst2}.}
\label{snips}
\end{center}
\end{table*}

\subsection{Experimental Setup}
\bigskip

In the first task, we use the SST-2 (Stanford Sentiment Treebank, version 2) dataset \cite{SocherEtAl2013:RNTN} to conduct sentiment analysis experiments. SST-2 is a sentence binary classification dataset with train/dev/test splits provided and two types of sentence labels, i.e., positive and negative.

The second task involves two essential problems in SLU, which are intent classification (IC) and slot labeling (SL). In IC, the model needs to detect the intention of a text input (i.e., utterance, conveys). 
For example, for an input of \emph{I want to book a flight to Seattle}, the intention is to book a flight ticket, hence the intent class is \emph{bookFlight}. 
In SL, the model needs to extract the semantic entities that are related to the intent. 
From the same example, \emph{Seattle} is a slot value related to booking the flight, i.e., the destination. 
Here we experiment with the Snips dataset \cite{coucke2018snips}, which is widely used in SLU research. 
This dataset contains test spoken utterances (text) classified into one of 7 intents.

In both tasks, we used the open-sourced GluonNLP BERT model to perform text classification. 
For evaluation, sentiment analysis is measured in accuracy, whereas IC and SL are measured in accuracy and F1 score, respectively. 
BERT is fine-tuned on train/dev sets and evaluated on test sets.

We down-sampled SST-2 and Snips training sets from $100\%$ to $10\%$ with intervals being $10\%$. 
BERT's performance is reported for each down-sampled setting in Table \ref{sst2} and Table \ref{snips}. 
We used entire test sets for all model evaluations. 

To compare, we compute the proposed data metrics, i.e., diversity, density, and homogeneity, on the original and the down-sampled training sets.

\subsection{Experimental Results}
\bigskip
We will discuss the three proposed characteristic metrics, i.e., diversity, density, and homogeneity, and model performance scores from down-sampling experiments on the two public benchmark datasets, in the following subsections:

\subsubsection{SST-2}
\bigskip
In Table \ref{sst2}, the sentiment classification accuracy is $92.66\%$ without down-sampling, which is consistent with the reported GluonNLP BERT model performance on SST-2. 
It also indicates SST-2 training data are differentiable between label classes, i.e., from the positive class to the negative class, which satisfies our assumption for the characteristic metrics.

Decreasing the training set size does not reduce performance until it is randomly down-sampled to only $20\%$ of the original size. 
Meanwhile, density and homogeneity metrics also decrease significantly (highlighted  in  bold in Table \ref{sst2}), implying a clear relationship between these metrics and model performance.

\subsubsection{Snips}
\bigskip
In Table \ref{snips}, the Snips dataset seems to be distinct between IC/SL classes since the IC accurcy and SL F1 are as high as $98.71\%$ and $96.06\%$ without down-sampling, respectively. 
Similar to SST-2, this implies that Snips training data should also support the inter-class differentiability assumption for our proposed characteristic metrics.

IC accuracy on Snips remains higher than $98\%$ until we down-sample the training set to $20\%$ of the original size. 
In contrast, SL F1 score is more sensitive to the down-sampling of the training set, as it starts decreasing when down-sampling. 
When the training set is only $10\%$ left, SL F1 score drops to $87.20\%$. 

The diversity metric does not decrease immediately until the training set equals to or is less than $40\%$ of the original set. 
This implies that random sampling does not impact the diversity, if the sampling rate is greater than $40\%$. 
The training set is very likely to contain redundant information in terms of text diversity. 
This is supported by what we observed as model has consistently high IC/SL performances between $40\%$-$100\%$ down-sampling ratios. 

Moreover, the biggest drop of density and homogeneity (highlighted in bold in Table \ref{snips}) highly correlates with the biggest IC/SL drop, at the point the training set size is reduced from $20\%$ to $10\%$. 
This suggests that our proposed metrics can be used as a good indicator of model performance and for characterizing text datasets.

\section{Analysis}
\bigskip
We calculate and show in Table \ref{corr} the Pearson's correlations between the three proposed characteristic metrics, i.e., diversity, density, and homogeneity, and model performance scores from down-sampling experiments in Table \ref{sst2} and Table \ref{snips}. 
Correlations higher than $0.5$ are highlighted in bold. 
As mentioned before, model performance is highly correlated with density and homogeneity, both are computed on the train set. 
Diversity is only correlated with Snips SL F1 score at a moderate level.

\bigskip
\begin{table}[!h]
\begin{center}
\begin{tabularx}{\columnwidth}{|l|X|X|X|}
      \hline
      Dataset&SST-2&Snips&Snips\\
      \hline
      Task Evaluation Metrics&Acc.&IC Acc.&SL F1\\
      \hline
      Corr. to Diversity&0.196&0.196&\bf{0.555}\\
      \hline
      Corr. to Density&\bf{0.637}&\bf{0.637}&\bf{0.716}\\
      \hline
      Corr. to Homogenity&\bf{0.716}&\bf{0.958}&\bf{0.983}\\
      \hline
\end{tabularx}
\caption{The Pearson's correlation (\emph{Corr.}) between proposed characteristic metrics (diversity, density, and homogeneity) and model accuracy (\emph{Acc.}) or F1 scores from down-sampling experiments in Table \ref{sst2} and Table \ref{snips}.}.
\label{corr}
\end{center}
\end{table}

These are consistent with our simulation results, which shows that random sampling of a dataset does not necessarily affect the diversity but can reduce the density and marginally homogeneity due to the decreasing of data points in the embedding space. 
However, the simultaneous huge drops of model performance, density, and homogeneity imply that there is only limited redundancy and more informative data points are being thrown away when down-sampling.
Moreover, results also suggest that model performance on text classification tasks corresponds not only with data diversity but also with training data density and homogeneity as well.


\section{Conclusions}
\bigskip

In this work, we proposed several characteristic metrics to describe the diversity, density, and homogeneity of text collections without using any labels.
Pre-trained language embeddings are used to efficiently characterize text datasets. Simulation and experiments showed that our intrinsic metrics are robust and highly correlated with model performance on different text classification tasks. 
We would like to apply the diversity, density, and homogeneity metrics for text data augmentation and selection in a semi-supervised manner as our future work.

\section{Bibliographical References}
\label{main:ref}



\end{document}